\documentclass[twoside,11pt]{article}

\usepackage{jmlr2e}

\usepackage[square,numbers]{natbib}
\usepackage{enumitem}
\usepackage{algpseudocode}
\newcommand{\lib}[1]{\textsc{DeepAL}}

\firstpageno{1}

\begin{document}

\title{DeepAL: Deep Active Learning in Python}

\author{\name Kuan-Hao Huang \email khhuang@cs.ucla.edu \\
       \addr Computer Science Department\\
       University of California, Los Angeles}

\maketitle

\begin{abstract}
We present \lib{}, a Python library that implements several common strategies for active learning, with a particular emphasis on deep active learning.
\lib{} provides a simple and unified framework based on PyTorch that allows users to easily load custom datasets, build custom data handlers, and design custom strategies without much modification of codes.
\lib{} is open-source on Github\footnote{\url{https://github.com/ej0cl6/deep-active-learning}} and welcome any contribution.
\end{abstract}

\section{Introduction}

Active learning is a popular solution to reduce the expensive cost of labeling~\cite{Settles09survey}.
Due to its practical value, many libraries are presented to help people adopt active learning to their applications, such as JCLAL~\cite{Pupo16jclal} based on Java and libact~\cite{Yang17libact} based on Python.
They provide general frameworks for active learning and include some built-in query strategies. However, since they are developed in the earlier years, those libraries are designed for traditional learning approaches, such as support vector machine~\cite{Cortes95svm} and random forest~\cite{Breiman01rf}.
Given that deep learning has gradually become a standard learning approach recently, the demand for an active learning library that is capable of deep neural models is rising.
Motivated by this demand, we present \lib{}, a Python library that implements several common strategies for deep active learning.
\lib{} provides a simple and unified framework based on PyTorch for pool-based active learning~\cite{Settles09survey} and includes several modules that can be easily extended to custom datasets, custom data handlers, and custom query strategies.
We hope that \lib{} can be a useful tool for both practical applications and research purposes.

\section{Pool-Based Active Learning}

\lib{} Focuses on the pool-based active learning setting~\cite{Settles09survey}.
Given a labeled pool $\mathcal{D}_l = \{ (x_i, y_i) \}_{i=1}^{N_l}$ and an unlabeled pool $\mathcal{D}_u = \{ x_i \}_{i=1}^{N_u}$, a pool-based active learning algorithm first learns an initial classifier $f^{(0)}$ based on $\mathcal{D}_l$. 
Next, for each round $t = 1, 2, ..., T$, the algorithm selects $n$ examples from $\mathcal{D}_u$ according to a \emph{query strategy} and queries the corresponding labels of those selected examples. The $n$ examples are then moved from $\mathcal{D}_u$ to $\mathcal{D}_l$ and the algorithm learns a new classifier $f^{(t)}$ based the updated $\mathcal{D}_l$.
The objective is to design a good query strategy and make $f^{(0)}, f^{(1)}, f^{(2)}, ..., f^{(T)}$ perform well on the testing set $\mathcal{D}_{test} = \{ (x_i, y_i) \}_{i=1}^{M}$.

\section{\lib{}}

\lib{} consists of several modules to fit the scenario of pool-based active learning.

\paragraph{Data.}
The \texttt{Data} class maintains the labeled pool $\mathcal{D}_l$ and the unlabeled pool $\mathcal{D}_u$ as well as the testing set $\mathcal{D}_{test}$. 
Some important attributes are listed as follows.

\begin{itemize}[topsep=5pt, itemsep=0pt]
    \item \texttt{Data.X\_train} and \texttt{Data.Y\_train}: a list of examples and the corresponding labels in $\mathcal{D}_l \cup \mathcal{D}_u$.
    \item \texttt{Data.Y\_test} and \texttt{Data.Y\_test}: a list of examples and the corresponding labels in the testing set $\mathcal{D}_{test}$.
    \item \texttt{Data.labeled\_idxs}: a binary numpy array to indicate which examples in \texttt{Data.X\_train} and \texttt{Data.Y\_train} are labeled and which are not.
    \item \texttt{Data.handler}: a class inherits \texttt{torch.utils.data.Dataset} that pre-processes the data (\texttt{X} and \texttt{Y}) and convert them into Tensors. It is supposed to support two functions \texttt{Data.handler.\_\_getitem\_\_} and \texttt{Data.handler.\_\_len\_\_}.
\end{itemize}

\paragraph{Net.}
The \texttt{Net} class defines the architecture of classifier $f$ and the corresponding training parameters.
Some important attributes and methods are listed as follows.

\begin{itemize}[topsep=5pt, itemsep=0pt]
    \item \texttt{Net.net}: a class inherits \texttt{torch.nn.Module} that specifies the architecture of classifier~$f$. It is supposed to support \texttt{Net.net.forward} and \texttt{Net.net.get\_embedding\_dim}, where the latter function has to return the size of hidden representations.
    \item \texttt{Net.params}: a dictionary that contains all the parameters for training, such as the number of epochs, the batch size, and the learning rate.
    \item \texttt{Net.train(data)}: a function that specifies the training process. It trains a classifier  \texttt{Net.clf} from \texttt{data}.
    \item \texttt{Net.predict(data)}: a function that uses \texttt{Net.clf} to make predictions for \texttt{data} and returns the corresponding Tensors.
    \item \texttt{Net.predict\_prob(data)}: a function that uses \texttt{Net.clf} to make predictions with probabilities for \texttt{data} and returns the corresponding Tensors.
    \item \texttt{Net.get\_embeddings(data)}: a function that uses \texttt{Net.clf} to generate hidden representations for \texttt{data} and returns the corresponding Tensors.
\end{itemize}

\paragraph{Strategy.}
The \texttt{Strategy} class specifies the details of the query strategy.
Some important attributes and methods are listed as follows.

\begin{itemize}[topsep=5pt, itemsep=0pt]
    \item \texttt{Strategy.dataset}: a \texttt{Data} instance that maintains the labeled pool $\mathcal{D}_l$ and the unlabeled pool $\mathcal{D}_u$ as well as the testing set $\mathcal{D}_{test}$.
    \item \texttt{Strategy.net}: a \texttt{Net} instance that specifies the architecture of classifier $f$ and the corresponding training parameters.
    \item \texttt{Strategy.query(n)}: a function that implements the rule to select $n$ examples from the unlabeled pool $\mathcal{D}_u$.
    \item \texttt{Strategy.update(query\_idxs)}: a function that updates the labeled pool $\mathcal{D}_l$ and the unlabeled pool $\mathcal{D}_u$ with the given \texttt{query\_idxs}.
    \item \texttt{Strategy.train()}: a function that updates \texttt{Strategy.net} with the current labeled pool $\mathcal{D}_l$.
\end{itemize}
In \lib{}, we implement several common query strategies, including:

\begin{itemize}[topsep=5pt, itemsep=0pt]
    \item \textbf{Random sampling}: randomly select examples from $\mathcal{D}_u$.
    \item \textbf{Least confidence}~\cite{Lewis94least}: select examples with least confidence
    \[ x^* = \arg\max_x \, 1 - P(\hat{y}| x), \]
    where $\hat{y}$ is the most probable class label.
    \item \textbf{Margin sampling}~\cite{Scheffer01margin}: select examples with smallest margins
    \[ x^* = \arg\min_x \, P(\hat{y}_1| x)-P(\hat{y}_2| x), \]
    where $\hat{y}_1$ and $\hat{y}_2$ are the first and second most probable class labels.
    \item \textbf{Entropy sampling~}\cite{Settles09survey}: select examples with largest entropy
    \[ x^* = \arg\max_x \, - \sum_y P(y|x)\log P(y|x). \]
    \item \textbf{Uncertainty sampling with dropout estimation}~\cite{Gal17bald}: select examples with most uncertainties estimated by dropouts. We implement the above three uncertainties: least confidence, smallest margins, and largest entropy.
    \item \textbf{Bayesian active learning disagreement} (BALD)~\cite{Gal17bald}: select examples with largest mutual information between predictions and model posterior.
    \item \textbf{Core-set selection}~\cite{Sener18coreset}: select examples from the score-set based on $k$-means and $k$-medians algorithms.
    \item \textbf{Adversarial margin}~\cite{Ducoffe18adv}: select examples with smallest margins approximated by adversarial perturbations. We implement two adversarial methods: basic iterative method~\cite{Kurakin17bim} and DeepFool~\cite{Moosavi-Dezfooli16deepfool}.
\end{itemize}

\paragraph{Usage.}

The framework for the whole active learning process is shown by the following pseudo-code\footnote{Please refer to \url{https://github.com/ej0cl6/deep-active-learning/blob/master/demo.py} for more details.}.

\begin{figure}[!ht]
\centering
\begin{minipage}{.45\linewidth}
\begin{algorithmic}[1]
\State dataset = Data()
\State net = Net()
\State strategy = Strategy(dataset, net)
\For{$t = 1,2,3, ..., T$}
    \State query\_idxs = strategy.query(n)
    \State strategy.update(query\_idxs)
    \State strategy.train()
\EndFor
\end{algorithmic}
\end{minipage}
\end{figure}
\noindent As illustrated by line 4 to line 8, for each iteration, the strategy selects $n$ examples for querying, updates the labeled pool and the unlabeled pool, and train a classifier.

\section{Conclusion}
We present \lib{}, a Python library that implements several common strategies for deep active learning.
\lib{} provides a simple and unified framework and flexible modules that allow users to easily develop  custom active learning strategies. We hope that \lib{} can be a useful tool for both practical applications and research purposes.

\bibliographystyle{plainnat}
\bibliography{cite}

\end{document}